\documentclass[letterpaper, 10 pt, conference]{ieeeconf}
\IEEEoverridecommandlockouts
\usepackage{cite}
\usepackage{amsmath,amssymb,amsfonts}
\usepackage{algorithm}
\usepackage[noend]{algpseudocode}
\usepackage{graphicx}
\usepackage{textcomp}
\usepackage{xspace}
\usepackage{censor}

\def\BibTeX{{\rm B\kern-.05em{\sc i\kern-.025em b}\kern-.08em
    T\kern-.1667em\lower.7ex\hbox{E}\kern-.125emX}}

\usepackage[dvipsnames]{xcolor} 

\newcommand{\comment}[3]{{\color{#1}\textbf{#2} #3}}
\newcommand{\evanc}[1]{\comment{Purple}{Evan C:}{#1}}
\newcommand{\davidc}[1]{\comment{Orange}{David C:}{#1}}
\newcommand{\laura}[1]{\comment{BlueGreen}{Laura:}{#1}} 
\newcommand{\mak}[1]{\comment{LimeGreen}{Mak:}{#1}}

\renewcommand{\comment}[3]{}

\begin{document}
\StopCensoring

\title{\LARGE \bf POrTAL: Plan-Orchestrated Tree Assembly for Lookahead* 
}


\author{\censor{Evan Conway**, David Porfirio**, David H. Chan, Mark Roberts** and Laura M. Hiatt}%
\blackout{\thanks{*This work was supported by the US Naval Research Laboratory}%
\thanks{Evan Conway is with the University of Virginia,
Charlottesville, Virginia, USA {\tt\small auj4kq@virginia.edu}}%
\thanks{David Porfirio is with George Mason University, Fairfax, Virginia, USA
{\tt\small dporfiri@gmu.edu}}%
\thanks{David H. Chan and Laura M. Hiatt are with the Navy
Center for Applied Research in AI, US Naval Research Laboratory,
Washington, D.C., USA
{\tt\small david.h.chan4.civ@us.navy.mil,} 
{\tt\small laura.m.hiatt.civ@us.navy.mil}}%
\thanks{Mark Roberts is with Iconium Labs, Tempe, AZ, USA
{\tt\small makro@iconiumlabs.com}}%
\thanks{**Work performed while at the US Naval Research Laboratory}%
}}

\newcommand{\alg}[0]{{\textit{POrTAL}}}
\newcommand{\ProcName}[1]{\textsc{#1}}
\newcommand{\eg}[0]{{e.g.,}}
\newcommand{\ie}[0]{{i.e.,}}

\maketitle
\thispagestyle{empty}
\pagestyle{empty}

\begin{abstract}
When tasking robots in partially observable environments, these robots must \textit{efficiently} and \textit{robustly} plan to achieve task goals under uncertainty.
Although many probabilistic planning algorithms exist for this purpose, these algorithms can be inefficient if executed with the robot's limited computational resources, or may produce policies that take more steps than expected to achieve the goal. 
We therefore created a new, lightweight, probabilistic planning algorithm, \textit{Plan-Orchestrated Tree Assembly for Lookahead (}\alg{}\textit{)}, that combines the strengths of two baseline planning algorithms, FF-Replan and POMCP.
We demonstrate that \alg{} is an anytime algorithm that  generally outperforms these baselines in terms of the final executed plan length given bounded computation time, especially for problems with only moderate levels of uncertainty.
\laura{ideally we say here that we perform better than FF-Replan and asymptotically approach POMCP} \evanc{(asymptotically approach with modifications)}
\end{abstract}




\section{INTRODUCTION}

Planning and acting under uncertainty remains an important, but computationally difficult problem to solve. This is especially true for robots, which pursue long-horizon goals in unpredictable environments. 
One common source of uncertainty is partial observability, where robots have only partial knowledge of their surrounding environment. Such problems are commonly formulated as partially observable Markov decision processes (POMDPs), which can be very expensive to solve, though there are several existing planning approaches that aim to solve such problems in a scalable way. FF-Replan \cite{yoon2007ff}, for example, quickly plans (and replans) based on a determinized version of the world. FF-Replan benefits from being able to quickly find plans to execute, but is not optimal and its performance often degrades as uncertainty increases \cite{little2007probabilistic}.  Partially Observable Monte Carlo Planning (POMCP) \cite{NIPS2010_edfbe1af}, by contrast, is an anytime planner that enables the robot to continuously improve its plan while it executes its task. POMCP is provably optimal given sufficient planning time, and so can be well suited to avoid dead-ends or other errors in highly uncertain domains. However, POMCP requires the construction of large search trees; finding good solutions, even if not optimal, can therefore be computationally intensive. This limitation is especially apparent if the reward signal is weak, such as if reward is achieved only upon completion of the task. 



\begin{figure}[t]
  \includegraphics[width=\columnwidth]{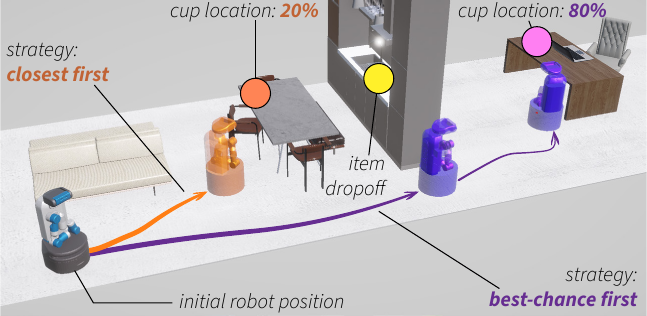}
  \caption{Office environment in which a robot is tasked to deliver a cup to the kitchen. The cup has a 20\% chance of being near the kitchen already and an 80\% chance of being on the desk. Although there is a smaller chance of the cup being in the kitchen, minimizing the number of anticipated steps to deliver the cup may require the robot to check near the kitchen first.}
  \label{fig:teaser}
\end{figure}

In this paper, we develop an approach that strikes a middle ground between FF-Replan and POMCP, providing rapid, strong solutions for domains with \textit{medium uncertainty}, namely those with known environment layouts but some hidden parameters, such as the locations of task-critical objects. Examples of medium-uncertainty domains include source localization during disaster response \cite{rhodes2023autonomous}, search and rescue \cite{khanal2025learning,chiou2022towards}, and household search \cite{paudel2025deploymenttime, ramrakhya2022habitat}. To illustrate, consider a human user in the office environment shown in Figure \ref{fig:teaser} who makes a delivery request to the robot---\textit{bring the cup to the kitchen}---which then autonomously plans and acts in the world to achieve the request. In this medium-uncertainty domain, the layout of the office is known to the robot due to previous exploration, but the cup's location is neither known to the human nor the robot. Instead, the robot possesses only a distribution of where the cup has been seen previously. 
Deciding which location to search first involves minimizing the \textit{expected} number of steps to achieve the goal, including the cost of potential backtracking. 


To better handle medium-uncertainty domains, we created a new anytime planning algorithm, \textit{Plan-Orchestrated Tree Assembly for Lookahead (}\alg{}\textit{)}.
\alg{} works by combining the strengths of both POMCP and FF-Replan.
Like POMCP, \alg{} iteratively generates a search tree of possible future actions and observations. Unlike POMCP, \alg{} utilizes a classical planner to limit the breadth-wise expansion of the tree and more rapidly facilitate its depth-wise expansion towards the goal. 
Specifically, when expanding a node in the search tree, \alg{} draws inspiration from FF-Replan by sampling a determinized variant of the problem, using a classical planner to generate a plan, and inserting the entire plan into the tree at once as a new branch. 
By repeating this process, \alg{} iteratively builds a search tree of many possible futures. \laura{still confusing where branches happen, vs. just having a sequence of full plans coming from the root -- maybe it assumes more knowledge then I currently have of POMCP but would be nice to address if possible.} To guide this search, \alg{} labels tree nodes as \textit{meaningful} if they represent critical junctures where an observation could invalidate the determinized plan's assumptions. These meaningful nodes are then prioritized for further exploration. 

We experimentally compared \alg{} to POMCP and FF-Replan and found that for the medium-uncertainty scenarios that we are interested in, \alg{} generates higher-quality solutions than FF-Replan, whose strategy of repeatedly committing to and replanning from the most likely state often gives poor results. Furthermore, our anytime performance analysis reveals that \alg{} converges to these solutions significantly faster than POMCP. While POMCP has theoretical guarantees of eventually converging to an optimal solution, its broad initial exploration is ineffective  under the time constraints typical of robotics applications. \alg{}, by contrast, provides a practical trade-off, finding good, though not necessarily optimal, solutions within a short time budget. 


\laura{TODO / the plan: add part about always beating FF-Replan etc. should be abstract, here, discussion and conclusion. ALSO should be in results section with experimental results confirming it. we will prove in the next couple days for the final version of submission.}
Our contributions are as follows:
\begin{itemize}
  \item \textit{Technical}---The \alg{} algorithm, which balances computational efficiency and solution quality for medium-uncertainty domains. 
  \item \textit{Empirical}---An evaluation of \alg{}, including comparisons against FF-Replan and POMCP baselines.
\end{itemize}
\section{RELATED WORK}
We are interested in planning in medium-uncertainty situations, such as planning for robots in households \cite{paudel2025deploymenttime} or search-and-rescue scenarios \cite{khanal2025learning} in which the locations of task-critical entities are unknown. 
In these cases, the robot operates with a pre-existing map but must reason about where to find pertinent people or items. 
As stated above, one common formulation for these types of problems is the partially observable Markov decision process (POMDP), which models a Markov decision process in which agents do not know exactly what state they are in, but instead track {\em beliefs} over states they may be in \cite{KAELBLING199899}. POMDPs, however, can be very expensive to solve; a large body of work focuses on improving the efficiency of finding high-quality solutions. 

One such algorithm is QMDP \cite{littman1995learning}, which simplifies POMDPs by assuming full observability after one step. \mbox{QMDP-Net} encodes QMDP within a neural network architecture \cite{karkus2017qmdp}, and so represents a class of methods that embed planning algorithms directly into differentiable neural architectures, further improving efficiency. 
In contrast to \mbox{QMDP-Net}, we are interested in developing anytime, online solutions to planning problems. 

DESPOT (Determinized Sparse Partially Observable Tree) is an anytime POMDP planning algorithm that balances search accuracy and efficiency by sampling a small set of representative scenarios from the belief distribution and constructing a sparse belief tree \cite{somani2013despot}. By using upper and lower bounds to prune the search space, DESPOT achieves strong performance guarantees while remaining computationally feasible for complex problems. 
\alg{} trades the formal guarantees of DESPOT for a more aggressive, heuristic-guided exploration by using each sampled state to generate a full determinized plan. This action sequence acts as a deep, goal-directed heuristic probe, avoiding the potentially expensive breadth-wise exploration of the action space that DESPOT must still contend with.  

POMHDP (Partially-Observable Multi-Heuristic Dynamic Programming) solves POMDPs using multiple heuristics to improve efficiency, while still asymptotically converging on the optimal answer \cite{kim2019pomhdp}. This is similar to what our approach does, which uses heuristics to guide tree search. 
However, POMHDP's heuristics require domain knowledge whereas \alg{}'s are domain independent. 

Monte Carlo Tree Search (MCTS) is a well-known algorithmic technique that has been used to manage the complexity of planning by using random sampling to reduce the number of solutions that need to be explored while still guiding the algorithm towards high-quality solutions \cite{coulom2006efficient}. A variant, POMCP, has been shown to be particularly useful in helping to scalably solve POMDPs \cite{NIPS2010_edfbe1af}, with additional improvements found by performing progressive widening \cite{sunberg2018online}. We use both of these aspects in our work, basing our approach on POMCP while gradually increasing the number of children under various tree nodes.


Curtis et al. \cite{curtis2023task}'s STRUG improves exploration efficiency for long-horizon tasks by combining deterministic planning with exploratory actions to eliminate task-irrelevant uncertainty. Our current evaluation demonstrates \alg{} on domains where the only uncertainty is with task-relevant entities; if task-irrelevant uncertainty were present as well, \alg{} could incorporate in the principles of STRUG to alleviate that complexity. 


\newcommand{\state}{\ensuremath{s}\xspace}
\newcommand{\states}{\ensuremath{S}\xspace}
\newcommand{\action}{\ensuremath{a}\xspace}
\newcommand{\actions}{\ensuremath{A}\xspace}
\newcommand{\observation}{\ensuremath{o}\xspace}
\newcommand{\observations}{\ensuremath{\Omega}\xspace}
\newcommand{\mdptransition}{\ensuremath{\mathcal{P}}\xspace}

\newcommand{\belief}{\ensuremath{b}\xspace}
\newcommand{\beliefs}{\ensuremath{B}\xspace}
\newcommand{\beliefsk}[1]{\ensuremath{B_{#1}}\xspace}
\newcommand{\beliefOverHistory}{\ensuremath{\mathcal{B}}\xspace}
\newcommand{\beliefApprox}{\ensuremath{\hat{\mathcal{B}}}\xspace} 
\newcommand{\beliefApproxOverHistory}{\ensuremath{\hat{\mathcal{B}}(h)}\xspace} 

\newcommand{\planner}{\textsc{Planner}\xspace} 

\newcommand{\simulator}[0]{\textsc{Sim}\xspace}  

\newcommand{\traverse}{\ProcName{Traverse}\xspace}  

\newcommand{\hao}{\ensuremath{hao}\xspace}
\newcommand{\history}{\ensuremath{h}\xspace}
\newcommand{\historySet}{\ensuremath{\mathcal{H}}\xspace}

\newcommand{\countSim}{\ensuremath{N_{Sim}}\xspace} 

\newcommand{\searchTree}{\ensuremath{Tree}\xspace}  

\newcommand{\tightpar}[1]{\vspace{2mm} \noindent \textbf{#1}}

\section{BACKGROUND}

\alg{} builds on two approaches for 
probabilistic planning, FF-Replan and POMCP. 
 We first provide some formal definitions, then describe the two approaches in more detail.

\subsection{Preliminaries}

\tightpar{POMDP.}
A partially observable Markov decision process (POMDP) is a tuple $(\states, \actions, \Omega, \mathcal{I}, \mdptransition, O, R, \gamma)$ where:
\begin{itemize}
    \item $\states$ is the state space;
    \item $\actions$ is the action space;
    \item $\observations$ is the observation space;
    \item $\mathcal{I}(\state): \states \to [0, 1]$ is the initial state probability;
    \item $\mdptransition(\state_{t + 1} \mid \state_t, \action_t): \states \times \states \times \actions \to [0, 1]$ is the state transition probability;\footnote{While the formal POMDP definition allows for the possibility of non-deterministic state transitions, in this paper we assume all such transitions are deterministic. We discuss this further in Section \S\ref{sec:discuss}.}
    \item $O(\observation_{t + 1} \mid s_{t + 1}, a_t): \observations \times \states \times \actions \to [0, 1]$ is the observation probability;
    \item $R(\state, \action): \states \times \actions \to \mathbb{R}$ is the reward function; and 
    \item $\gamma \in [0, 1)$ is the discount factor. 
\end{itemize}

\tightpar{Histories, Beliefs, and Particles.}
A \textit{history} is a sequence of actions and observations $h_t = \langle \action_0, \observation_0, \dots, \action_t, \observation_t\rangle$, or $h_ta_{t + 1} = \langle \action_0, \observation_0, \dots, \action_t, \observation_t, \action_{t + 1}\rangle$, where $\action_i \in \actions$ and $\observation_i \in \observations$. Let \historySet denote the set of all histories. For convenience, we may omit brackets or subscripts when the context is clear. 


Both POMCP and \alg{} represent the search space using a search tree, where nodes contain
histories. The root corresponds to the agent's current history (\ie{} the sequence of actions and observations received so far). Then, levels of the tree alternate between \textit{observation nodes}, which contain histories $h_t$ ending with an observation $o_t$, and \textit{action nodes}, which contain histories $h_ta_{t + 1}$ ending in an action $a_{t + 1}$. 

Because we assume partial observability with an unknown initial state, the agent maintains some \textit{belief} about the true state of the world. 
Let $\beliefOverHistory(\state \mid \history) : \states \times \historySet \to [0, 1]$ be the probability of state $\state$ given history $h$. That is, for $\state \in \states$ and $\history = a_0 o_0 \dots a_t o_t \in \historySet$, $\beliefOverHistory(\state \mid \history)$ is the probability that the true state of the world is $\state$ given initial state distribution $\mathcal{I}$ and observations $o_0, \dots, o_t$ for executing actions $a_0, \dots, a_t$, respectively. Note that $\beliefOverHistory(\state \mid \varnothing) = \mathcal{I}(s)$. 

In theory, the agent's belief can be updated exactly using Bayes' theorem.
However, in large state spaces, Bayes updates quickly become computationally infeasible. To plan efficiently, \cite{NIPS2010_edfbe1af} uses an unweighted particle filter to approximate the belief state. That is, rather than computing $\beliefOverHistory(\state \mid \history)$ exactly, we can approximate the belief state as a multiset of particles $\beliefApproxOverHistory \subseteq \states$, where each \emph{particle} $s \in \beliefApproxOverHistory$ is a sample state.
Then, with a multiset of $k$ particles for history $h$, we can approximate $\beliefOverHistory(\state \mid \history) \approx \frac{1}{k} \sum_{s' \in \beliefApproxOverHistory} \delta_{s s'}$ where $\delta_{ss'}$ is the Kronecker delta function. This eliminates the need for expensive Bayes updates or even an explicit model of the POMDP, and can be implemented efficiently using just a black box simulator for the environment. 

\subsection{Determinized Replanning Using a Classical Planner}\label{sec:ffreplan} 

FF-Replan solves MDP problems by planning over a determinized probabilistic domain,  typically by choosing the most likely determinization \cite{yoon2007ff}. It computes an optimal (\ie{} shortest path) plan for that determinization and executes the plan until failure, at which point it replans. The result is a branching plan. Due to its historical success as a good heuristic \cite{little2007probabilistic}, FF-Replan influences our design of \alg{}, specifically in the use of determinization and subsequent invocation of classical planning.

\tightpar{Limitations.}
FF-Replan suffers some notable issues. First, it is not anytime; \ie{} its solution does not improve with greater compute time. 
Second, if its determinizing assumptions do not hold, FF-Replan needs to replan from scratch based on the updated state of the world. This can cause inefficiencies in overall plan execution (e.g., backtracking). 
For example, in Figure~\ref{fig:teaser}, the best solution is for the robot to check for the cup on the kitchen table on its way to the desk; this eliminates backtracking if the cup is not on the desk. FF-Replan does not take multiple possibilities into account when planning, however, and so will generally first visit the desk, and then backtrack if the cup is not there. 

\subsection{Partially Observable Monte Carlo Planning (POMCP)}\label{sec:pomcp}



POMCP is a \textit{partially observable} variant of Monte Carlo Tree Search (MCTS) \cite{brown2012MCTS}.
As an anytime probabilistic planning algorithm for POMDPs, it iteratively constructs a search tree over all possible paths in a plan and explores different paths based on expected reward \cite{NIPS2010_edfbe1af}. 
Similar to MCTS, each iteration is comprised of a \textit{simulation} through the existing tree space, which ends in a \textit{rollout} through unexplored tree space. 
In contrast to MCTS, however, POMCP operates over belief states resulting from uncertainty in the environment and from the robot's actions.

Rather than planning to achieve a goal like FF-Replan, both MCTS and POMCP strive to achieve an optimal \textit{policy} over a reward space. Rewards received during rollouts are propagated up the tree, weighing certain paths in the tree more heavily than others. We incorporate aspects of POMCP into our design of \alg{}---namely the iterative construction of a search tree---because of its performance guarantees.

POMCP also uses particles to approximate the belief state instead of explicitly computing Bayes updates; this 
improves scaling for large belief states, and requires only a black-box simulator, \simulator. This simulator call $(s',o,r) \sim \simulator(s,a)$ takes in a state and an action, and outputs a new state, observation, and reward, according to the transition probabilities for the POMDP. This removes the need to explicitly compute the POMDP transition probabilities.

\tightpar{Limitations.}
Although POMCP converges to an optimal policy, it struggles along two dimensions. First, upon encountering a portion of the search tree that has not yet been explored (such as the root at the beginning of search) POMCP initially examines \textit{all} possible paths to a goal. Only after multiple simulations will more favorable paths be weighted more highly than others. As a result, POMCP can require substantial effort to narrow its search. This issue is compounded as the number of possible actions for the robot to perform at any given step increases, which increases the breadth of its search. Second, convergence can be slowed under sparse rewards, such as if reward is only received when the goal is achieved. In sufficiently complex problems with long planning horizons, few rollouts will receive any reward, drastically increasing the sample complexity.

\section{THE \alg{} ALGORITHM}\label{sec:alg}

Like POMCP, \alg{} uses a history tree representation, with particles at each node representing the belief. \alg{} uses an FF-Replan-like process to generate plans from various points in the tree, which are then added to the tree. This allows \alg{} to focus on promising sequences of actions instead of considering all actions, at the cost of potentially ignoring better actions.

\subsection{\alg{} Algorithm Walkthrough}




Algorithm~\ref{alg:main} depicts the key procedures of \alg{}. 
Procedure \ProcName{Search} (Lines~\ref{alg:main:start}--\ref{alg:main:end}) is the outer loop, and represents the robot \textit{searching} for the best action to perform next. 
Each search iteration begins by sampling a state $s \sim \beliefApproxOverHistory$ (Line~\ref{alg:main:sample}).  \traverse then finds a point from which to expand the tree and calls \ProcName{Rollout} to expand the tree and update the action values (Line~\ref{alg:main:traverse}). Procedure \ProcName{Timeout} (Line~\ref{alg:main:timeout}) represents the amount of time that the robot takes to complete an action, which we use here to define the amount of time that the robot is \textit{allowed} to plan per action. When the timeout is reached, the robot will select the action that has the maximum value at the moment when the time ran out. In practice, the actual timeout value may vary based on how long the robot takes to execute its current action while it plans for its next action. 

Within \traverse, \alg{} performs a check (Line \ref{alg:traverse:plan:decision}) at each observation node to see if it should expand the search tree from that node. If so, \alg{} uses a planner \planner on a specific state \state to obtain a plan $\langle a_0, \dots, a_n\rangle$ and then calls \ProcName{Rollout} on that plan (Lines~\ref{alg:traverse:plan:start}--\ref{alg:traverse:plan:end}) to add it to the search tree. Otherwise, \alg{} continues to move down the search tree by using UCT to select a new action and using \simulator to sample a resulting state and observation (Lines~\ref{alg:traverse:uct:start}--\ref{alg:traverse:end}), the same procedure as POMCP.

The choice of when to expand the search tree is determined by 
the set of {\em meaningful} nodes $\mathcal{M}$, and the $\ProcName{Expand}$ procedure (in Line~\ref{alg:traverse:plan:decision}). $\mathcal{M}$ is a subset of observation nodes that represent points where the plan has failed and replanning is required. $\ProcName{Expand}$ is a procedure that determines whether a node should be expanded with more actions. It checks if $\countSim(h) < k \cdot N(h)^{\alpha}$,  where $\countSim(h)$ is the number of rollouts performed from $h$, $N(h)$ is the number of node visits, and $k$ and $\alpha$ are hyperparameters. This formula, taken from the literature on progressive widening \cite{couetoux2011continuous}, balances between expanding repeatedly from the same nodes higher up in the tree and expanding from a more diverse array of nodes, including those lower down in the tree. However, other $\ProcName{Expand}$ functions could induce different behavior.

\begin{algorithm}
\caption{\alg{} }\label{alg:main}

\begin{algorithmic}[1]
\Procedure{\ProcName{Search}}{$h$} \label{alg:main:start}
    \While{not \ProcName{Timeout}$()$} \label{alg:main:timeout}
        \State $s \sim \beliefApproxOverHistory$ \label{alg:main:sample}
        \State $\traverse(s, h)$ \label{alg:main:traverse} \label{alg:main:end}
    \EndWhile
    \State \Return $\operatorname*{argmax}_{b} V(hb)$
\EndProcedure

\vspace{2mm}
\Procedure{\traverse}{$s, h$}  \label{alg:traverse:start}
    \State $N(h) \leftarrow N(h) + 1$
    \If{$h \in \mathcal{M}$ and $\ProcName{Expand}(N(h),\countSim(h))$} \label{alg:traverse:plan:decision}
        \State $\countSim(h) \leftarrow \countSim(h) + 1$ \label{alg:traverse:plan:start}
        \State $(a_0, \dots, a_n) \sim \planner(s)$ 
        \State \ProcName{Rollout}$(s, h, (a_0, \dots, a_n))$ \label{alg:traverse:plan:end}
    \Else
        \State $a \leftarrow \operatorname*{argmax}_{b} V(hb) + c \sqrt{\frac{\log N(h)}{N(hb)}}$ \label{alg:traverse:uct:start}
        \State $N(ha) \leftarrow N(ha) + 1$
        \State $(s', o, r) \sim \simulator(s, a)$ 

        \State $V(h a) \leftarrow V(h a) - \gamma \frac{|\beliefApprox(h a o)|}{|\beliefApprox(h)|} V(h a o)$ \label{alg:traverse:value_update:start}
        \State $\traverse(s', hao)$
        \State $V(h a) \leftarrow V(h a) + \gamma \frac{|\beliefApprox(h a o)|}{|\beliefApprox(h)|}  V(h a o)$

        \State $V(h) \leftarrow \operatorname*{max}_{b} V(hb)$ \label{alg:traverse:end}
    \EndIf
\EndProcedure

\vspace{2mm}
\Procedure{\ProcName{Rollout}}{$s, h, (a_k, \dots, a_n)$} \label{alg:rollout:start}
    \State $(s', o, r) \sim \simulator(s, a_k)$ \label{alg:rollout:simulate}
    
    \If{$h a_k \notin T$} \label{alg:rollout:expansion:start}
        \State $T(h a_k) \leftarrow \langle 0, 0 \rangle$ \label{alg:rollout:expansion:add_action}
        \For{$s_{alt} \in \beliefApproxOverHistory$} \label{alg:rollout:expansion:add_observations:start}
            \State $(s'_{alt}, o_{alt}, r_{alt}) \sim \simulator(s_{alt}, a_k)$ \label{alg:rollout:expansion:add_observations:simulate}
            \If{$h a_k o_{alt} \notin T$}
                \If{$o_{alt} \neq o$} \label{alg:rollout:expansion:add_observations:check_observation}
                    \State $\mathcal{M} \leftarrow \mathcal{M} \cup \{ h a_k o_{alt} \}$ \label{alg:rollout:expansion:add_observations:meaninful}
                \EndIf
                \State $T(h a_k o_{alt}) \leftarrow \langle \ProcName{Approx}(h a_k o_{alt}), \emptyset \rangle$

            \EndIf
            \State $\beliefApprox(h a_k o_{alt}) \leftarrow \beliefApprox(h a_k o_{alt}) \cup \{ s'_{alt} \}$ \label{alg:rollout:expansion:add_observations:belief_update}
            \State $V(h a_k) \leftarrow V(h a_k) + \frac{1}{|\beliefApprox(h)|} \cdot r_{alt}$
        \EndFor
    \EndIf \label{alg:rollout:expansion:end}

    \State $V(h a_k) \leftarrow V(h a_k) - \gamma \frac{|\beliefApprox(h a_k o)|}{|\beliefApprox(h)|}  V(h a_k o)$ \label{alg:rollout:value_update:start}
    \State $\ProcName{Rollout}(s', h a_k o, (a_{k+1}, \dots, a_n))$
    \State $V(h a_k) \leftarrow V(h a_k) + \gamma \frac{|\beliefApprox(h a_k o)|}{|\beliefApprox(h)|}  V(h a_k o)$
    \State $V(h) \leftarrow \operatorname*{max}_{b} V(hb)$  \label{alg:rollout:end}
\EndProcedure
\Statex
\end{algorithmic}
\end{algorithm}

\ProcName{Rollout} (Lines~\ref{alg:rollout:start}--\ref{alg:rollout:end}) adds the plan $\langle a_0, \dots, a_n\rangle$ generated by \planner to the search tree. At each step, if the action node for the next action $a_k$ is not in the search tree, then we add that node (Lines~\ref{alg:rollout:expansion:start}--\ref{alg:rollout:expansion:add_action}). \ProcName{Rollout} then initializes the following observation nodes by using \simulator to simulate the effect of applying that action to \textit{all} particles $s_{alt} \in \beliefApproxOverHistory$ at the current observation node (Lines~\ref{alg:rollout:expansion:add_observations:start}--\ref{alg:rollout:expansion:add_observations:simulate}). This gives an observation $o_{alt}$ and new state $s'_{alt}$, which is added to the relevant observation node to form the belief $\hat{\mathcal{B}}(h a_k o_{alt})$ at that node (Line~\ref{alg:rollout:expansion:add_observations:belief_update}).

\ProcName{Rollout} additionally checks the observation $o_{alt}$ against the observation $o$ for the particle that was initially planned from, $s$ (Lines~\ref{alg:rollout:simulate}, \ref{alg:rollout:expansion:add_observations:check_observation}). If $o_{alt} \neq o$, then the observation for that particle does not match with the observation expected from $s$. If this observation node is not part of the search tree $T$, \ProcName{Rollout} marks the new observation node as \textit{meaningful}, adding it to the set of meaningful nodes $\mathcal{M}$ (Line~\ref{alg:rollout:expansion:add_observations:meaninful}). One intuition for this is that if the observations match ($o_{alt} = o$), then \alg{} can continue following the plan $(a_0, ..., a_n)$. However, if the observations do not match ($o_{alt} \neq o$) then \alg{} doesn't know if it can continue following the plan, so it would like to try to generate new plans. Another intuition is that \textit{meaningful} nodes correspond to points where FF-Replan might have to replan.

While adding a plan in $\traverse$ and $\ProcName{Rollout}$, \alg{} simultaneously estimates the value of each node (Lines~\ref{alg:traverse:value_update:start}--\ref{alg:traverse:end}, \ref{alg:rollout:value_update:start}--\ref{alg:rollout:end}). The value of an observation node is the return of the best action available at that node, so $V(h) = \max_b V(hb)$. If an observation node has no children, \alg{} instead estimates its return using the approximation $V(h) = \ProcName{Approx}(h)$. We treat $\ProcName{Approx}$ as a black box, because different value approximation functions might be preferred in different cases. The value of an action node is the sum of the expected immediate reward and the discounted expected future return, $V(ha) = \mathbb{E}[r_t + \gamma \cdot G_t \mid a_t = a] = \mathbb{E}[r_t  \mid a_t = a] + \gamma \cdot \mathbb{E}_o[V(hao) \mid a_t = a]$.

When $\ProcName{Search}$ runs out of time, the action $a_t$ with the highest value $V(ha_t)$ is chosen as the next action to take. The agent then performs that action and receives a real observation $o_t$, with $T(h a_t o_t)$ becoming the new root node. 

\subsection{Key Differences From POMCP \& FF-Replan}

\tightpar{Key differences from POMCP.}
The key differences between \alg{} and POMCP lie in their exploration strategies. POMCP uses a broad exploration that expands its search tree via single-step action selection, then estimates node values with a computationally cheap but noisy random rollout. In contrast, \alg{} replaces the random rollout with a more computationally expensive classical planner that injects an entire plan as a deep branch that leads to a goal. This heuristic approach, while more expensive per simulation, generates stronger reward signals. 

Additionally, while POMCP relies solely on the UCB1 formula to traverse the tree and select nodes to expand, \alg{} focuses its search by prioritizing {\em meaningful} nodes---points where a determinized plan's assumptions may diverge from observations. This targeted exploration focuses on resolving key uncertainties about the environment. This is a strong, practical choice for the medium-uncertainty domains, but notably \alg{} trades the asymptotic optimality guarantees of POMCP in favor of greater anytime performance.

\tightpar{Key differences from FF-Replan.} The key differences between \alg{} and FF-Replan stem from how they handle uncertainty and explore future possibilities. 
FF-Replan employs a reactive, greedy strategy: it generates a single plan from the most probable state without trying to examine the quality of that plan, and then follows it until an unexpected observation forces a failure. In contrast, \alg{} generates \textit{multiple} plans by sampling states from the current belief distribution. This allows it to find a variety of possible actions to take, and weigh different outcomes to find a more robust policy. 
By sampling the 
state to plan from instead of only choosing the most probable state, it
avoids pathological scenarios where the most probable state corresponds to a plan that works poorly in general.


\subsection{\alg{} Implementation Details}
\label{sec:implementation}
Our specific implementation of \alg{} makes several important design decisions.
The known aspects of the world (e.g., environment, items, people), including the robot's goal, are initially specified in the Planning Domain Definition Language (PDDL) \cite{fox2003pddl2}. Uncertain aspects of the world, such as unknown locations of items/people, are given to the robot as probability distributions (see Section \S\ref{subsec:uncert}); once these locations are resolved, they are updated in the determinized representation. We used Fast Downward as the classical planner for \alg{} \cite{helmert2006fast}. 

\alg{} gets a constant reward of $-1.0$ for each action taken, with no discounting. This reward function corresponds to minimizing the expected number of actions needed to achieve the goal. 
An exploration value of $c = 20$ is used. 
For the $\ProcName{Expand}$ procedure, $\countSim(h) < k \cdot N(h)^{\alpha}$, \alg{} uses $k = \frac{1}{2}$ and $\alpha = 1$. These values were determined empirically.
In our implementation of $\ProcName{Approx}$, nodes' values are approximated as the mean value of their neighbors within the tree, or zero if they have no neighbors.

\section{EVALUATION}

\begin{figure}[t]
  \includegraphics[width=\columnwidth]{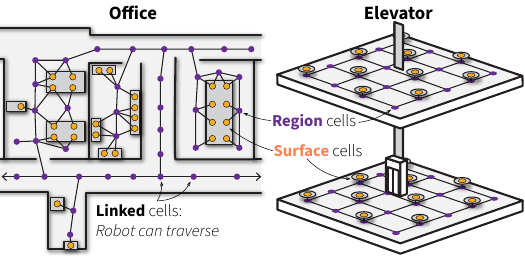}
  \caption{A visual depiction of the domains used in our evaluation. The \textit{office} domain (left) has both rooms and hallways (white) with tables (grey). Purple nodes are region cells and orange nodes are surface cells. Adjacent cells are linked based on the robot's ability to travel between them. The \textit{elevator} domain consists of two floors of a building linked by an elevator. Each floor is divided into a 5x4 grid of region cells that the robot can travel between. Some region cells have surface cells connected to them.}
  \label{fig:domains}
\end{figure}

We evaluate \alg{} across two different domains and with varying levels of uncertainty. Given these conditions, we perform two different experiments: one comparing \alg{} to baselines, and one further exploring the performance of \alg{} under different parameters. We describe these next.

\subsection{Domain}
We tested \alg{}'s performance in two separate planning domains---an \textit{office} domain, and a multi-floor \textit{elevator} domain. Each domain supports \textit{cells}, \textit{items}, and \textit{people}. Cells are discrete spatial points used to discretize the environments. To facilitate clarity in our discussion, we further denote cells as \textit{regions}, denoting places in rooms/hallways/etc., or \textit{surfaces}, denoting furniture in the environment that objects can naturally be placed on (\eg{} tables). All cells are connected in a graph-like structure (see Fig. \ref{fig:domains}). The robot uses a \textit{move} action to traverse between connected cells. 

Items and people can be located at any cell. At any time, a robot can use a \textit{sense} action to detect what items or people (if any) are in its current cell.  Items can be grabbed and placed at any cell, given to and received from people; under certain circumstances they can also be placed inside another item such as a box. The robot can carry only one item at a time (items containing other items count as one). For the purposes of these domains, we assume people do not move.



\subsubsection{Office} The \textit{office} domain represents a workplace environment, including rooms, hallways, and tables in each room. Figure \ref{fig:domains} (top) shows its representation as region and surface cells. 
The robot's goal is to place both a \textit{cup} and a \textit{plate} in a \textit{box}. The initial locations of all three items are uncertain; the parameters of their uncertainty are discussed in \S\ref{subsec:uncert}. 

The primary source of complexity is that the goal can be achieved via multiple distinct trajectories, a trait of ``probabilistically interesting'' domains as defined by \cite{little2007probabilistic}. For example, the robot can find the cup, place the cup in the box, carry the box to the plate, and then also place the plate in the box. Alternatively, the robot could find the box first and then locate both items while carrying the box. Overall, this means the robot has to consider which items are best to search for earlier on in its task given the uncertainty in the world state. 

\subsubsection{Elevator} 

The \textit{elevator} domain consists of two identical floors joined by an elevator, with tables scattered on each floor. Figure \ref{fig:domains} (bottom) shows its representation as region and surface cells. We model the elevator as 10 sequential cells to capture the cost of switching floors. The robot's goal in this domain is to fetch a \textit{package}, deliver it to the \textit{recipient}, and then report to a \textit{staff member}. The initial locations of the \textit{package}, \textit{recipient}, and \textit{staff member} are all uncertain; the parameters of their uncertainty are discussed in \S \ref{subsec:uncert}.


The primary source of complexity in this domain arises from the robot needing to choose when to take the elevator to a different floor as opposed to staying on the same floor. Since riding the elevator costs 10 actions each way, traversing to the other floor is costly. The robot needs to be strategic in terms of when it should take the elevator in order to minimize plan length. This domain was constructed to test the algorithm’s ability to reason about costly commitments and long-range action dependencies, similar in spirit to some synthetic planning benchmark domains designed to expose limitations of particular planners (e.g., Triangle Tireworld in \cite{little2007probabilistic}). It also exhibits two key structural properties emphasized in that work: multiple distinct trajectories from the initial state to the goal, and mutual exclusion between competing courses of action due to the high penalty for changing floors. 

\subsection{Uncertainty Level}\label{subsec:uncert}
As discussed, we are interested in planning for robots operating in medium-uncertainty domains. To emulate this in our experiments, we used a two-step process to 
generate the initial distributions for each item/person whose location was unknown: We randomly selected a set of possible locations for each item/person, then 
assigned each possible location a probability that the associated item/person is there. 

As part of the evaluation, we varied the parameters of both steps of this process. Specifically, we explored different numbers of possible locations per item/person. We also tested three different distributions over locations: a uniform \textit{high variance} distribution, where each location has an equal chance of the item being there; a gradual \textit{medium variance} distribution, where each location ordered by increasing probability has 75\% of the probability of an item being at that location as the next location; and a sharper \textit{low variance} distribution, where each location has 50\% of the probability as the next location of an item being at that location. For example, the \textit{low variance} distribution with four possible locations for each item would have location probabilities of $\frac{1}{15}, \frac{2}{15}, \frac{4}{15},$ and $\frac{8}{15}$.

\subsection{Experimental Comparisons}
We evaluate \alg{} via (1) comparing its performance to two baselines, FF-Replan and POMCP (\textit{Algorithm Comparison}); and (2) comparing its performance to itself across several different \textit{timeout} values (\textit{Time Comparison}).  

\subsubsection{Algorithm Comparison}
The \textit{Algorithm Comparison} compares \alg{} to POMCP and FF-Replan baselines. We ran this comparison across both experimental domains. For these experiments, we consider two parameterizations each of \alg{} and POMCP: one in which the robot is given four seconds to plan per action, and another in which the robot is given 16 seconds to plan per action.
All algorithms are allotted five times as much planning time (a timeout of 20 seconds rather than 4 seconds, or 80 seconds instead of 16 seconds) at the very start to represent startup cost, in which an initial tree is constructed before the robot takes any actions; note that FF-Replan takes negligible time to plan. 

Preliminary evaluations indicated that using a flat reward function ($-1.0$ per action; see \S\ref{sec:implementation}) leads to sparse reward signals for POMCP, which can significantly slow learning. To ensure that POMCP is a viable baseline, we incorporate reward shaping. Specifically, POMCP is given a reward of $1.0$ for: (1) successfully finding an uncertain item; or (2) achieving manually-identified domain-specific subgoals, which in the office domain is placing the cup or plate into the box, and in the elevator domain is giving the package to the recipient or reporting to the staff member. These subgoals serve to guide the planner by focusing search in a manner reminiscent of landmark-guided classical planning techniques \cite{richter2010lama}. POMCP is also given a reward of $0.1$ for each {\em sense} action taken by the robot in order to encourage exploration. In summary, this reward shaping constitutes domain-dependent guidance, which we found necessary to ensure that POMCP converges to a useful policy in a reasonable timeframe.

A discount factor of $\gamma = 0.97$ is used along with $\epsilon = 0.01$. This gives a search horizon of around 150 steps (maximum search depth, including rollouts). 
The exploration constant $c = 0.1$ was experimentally determined to yield good results. 
We also vary the number of possible locations per item/person in $\{2, 4, 6, 8, 10\}$ and consider 
all three uncertainty variance levels (low, medium, and high) when assigning probabilities to these locations. 

\subsubsection{Time Comparison}
The \textit{Time Comparison} characterizes how \alg{} performs if given increasing timeout thresholds (see \S\ref{sec:alg}), ranging in $\{2, 4, 8, 16, 32\}$ 
seconds per action. 
For this experiment, we consider $4$ and $8$ candidate locations per item/person to conduct this comparison with under mid-low and a mid-high numbers of items, with specific probabilities assigned according to the high-variance distribution. 

\subsection{Results}

\begin{figure*}[!th]
  \includegraphics[width=\textwidth]{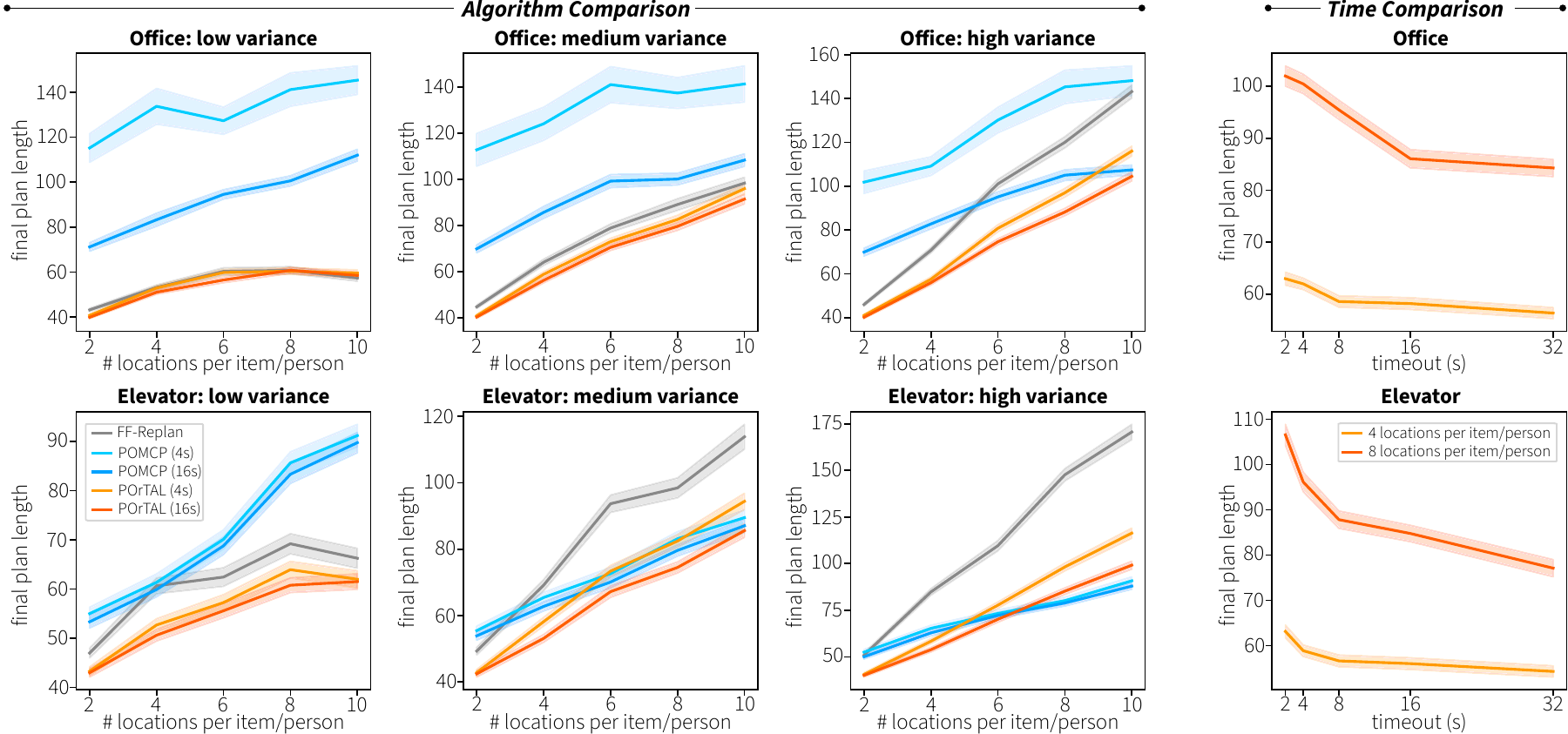}
  \caption{The results of our \textit{Algorithm Comparison} (left) and \textit{Time Comparison} (right). Error bands represent standard error of the mean. Lower is better.} 
  \label{fig:results}
\end{figure*}

Figure \ref{fig:results} depicts our results for the \textit{Algorithm} (left) and \textit{Time} (right) comparisons. Performance is measured by 
the number of steps 
to reach the goal (lower is better). 
Notably, this comparison favors POMCP through its domain-specific reward shaping, whereas \alg{} is entirely domain-independent. Despite POMCP having this additional information, \alg{} still achieves comparable or superior performance across most settings. 




\paragraph{\alg{} \textbf{Often} Works Well with Bounded Uncertainty}
The \textit{Algorithm Comparison} (Figure \ref{fig:results}, left and center) shows that \alg{} more strongly outperforms POMCP under the lower-variance distributions and less strongly outperforms---even sometimes underperforms---POMCP under more high variance distributions. Even so, for the high variance distribution, POMCP only begins to outperform \alg{} as the amount of uncertainty in the environment increases. \textbf{Interpretation:} We draw two conclusions from this result: First, \alg{} is more reliable than POMCP at producing good plans when there are fewer candidate locations. Second, \alg{} is better at producing plans in lower-variance environments. Overall, \alg{} outperforms POMCP when there is less uncertainty,
despite the additional domain-dependent reward shaping in POMCP. 


\alg{} performs consistently better than FF-Replan in high-variance distributions and at least as well as FF-Replan in low-variance distributions. This performance gap is particularly apparent in the elevator domain, where \alg{} significantly outperforms FF-Replan in all cases. \textbf{Interpretation:} We believe this result is because \alg{} is less prone to oscillating between different candidate locations when searching for objects, a behavior that FF-Replan exhibits. For example, if the most likely state in the elevator domain places an object on another floor, FF-Replan's greedy strategy would commit to taking the elevator immediately. If that object is not found, and its next most likely location is back on the original floor, FF-Replan would immediately take the elevator back. This costly oscillating behavior is avoided by \alg{}'s more robust sampling strategy, which can more accurately weigh the high cost of travel against the probability of finding the object.

\paragraph{\alg{} is Lightweight}
The 4-second variant of \alg{} outperforms the 16-second variant of POMCP in both office and elevator \textit{low variance} distributions, the office \textit{medium variance} distribution, and a substantial portion of the office \textit{high variance} distribution. In these scenarios, \alg{} requires less time to achieve better results than POMCP. \textbf{Interpretation:} \alg{} is lightweight, often requiring less time to produce better results than POMCP.

\paragraph{Planner Performance Improves with Time}
In the \textit{Algorithm Comparison} (Figure \ref{fig:results}, left), the 16-second variant of \alg{} and POMCP perform either better than or equal to their 4-second counterparts. The \textit{Time Comparison} (Figure \ref{fig:results}, right) similarly indicates that \alg{} improves with greater time allocated to the robot per action, though with diminishing returns after 10 to 20 seconds pass. \textbf{Interpretation:} While greater performance over time is a known phenomenon with POMCP, these results enable us to similarly conclude that \alg{} is an effective anytime algorithm. 

\laura{POrTAL always equaling or beating FF-REPLAN with enough time ALSO should be in results section with experimental results confirming it.}

\section{DISCUSSION}
\label{sec:discuss}

\laura{re-iterate heavily medium-uncertainty and anytime nature.}

\alg{} shows promise for contexts that require high-performing, anytime, and lightweight probabilistic planning solutions. Our results show that \alg{} improves upon POMCP for medium-uncertainty goal-directed problems by integrating classical planning with probabilistic search to create a focused exploration strategy. Whereas POMCP explores the belief space via single-step action sampling, \alg{} injects entire plans as deeper probes into the Monte Carlo search tree. This provides heuristic guidance by focusing search along complete action sequences known to reach the goal under specific determinizations, in contrast to the POMCP's broad exploration strategy. This is particularly helpful in domains without natural reward shaping, where the random rollouts of POMCP are unlikely to generate reward signals. By constructing plans known to reach the goal, \alg{} is able to receive reward signals more quickly, allowing the search to learn and prioritize more promising actions. This leads to improved performance in anytime planning where time constraints favor quick, lightweight solutions over slow convergence to an optimal one. 

Our POMCP baseline also incorporates domain-specific reward shaping via task-relevant subgoals, whereas \alg{} uses no domain-dependent guidance and instead relies on determinized classical plans. Despite this domain-knowledge advantage for POMCP, \alg{} performs comparably or better across most settings.

\alg{} also offers a more robust approach to handling domains with medium uncertainty than classical replanning methods like FF-Replan. In such domains, FF-Replan's strategy of committing to a plan based on the most-likely state results in a greedy strategy which frequently backtracks when new observations invalidate old assumptions. In contrast, \alg{} often produces higher-quality solutions by sampling its belief before each simulation. This sampling handles the constrained types of uncertainty typical of robotics domains, ensuring that the search evaluates determinized plans from many plausible world states, weighted by their likelihood. 



\subsection{Limitations and Future Work}
Our research has a few notable limitations to explore in future work. First, we have not yet tested \alg{} on a robot platform in a real-world domain. Testing onboard a robot would help create realistic values for our testing configuration, including how much time the robot has available to plan per action, realistic levels of uncertainty in different environments, and how this uncertainty is distributed. 

Relatedly, we make domain assumptions that may not be realistic for physical robots. For example, our planning approach uses dedicated \textit{sense} actions to resolve uncertainty. While this is necessary in some robot deployments to detect task-relevant objects, in others, the robot might continuously observe new information passively. 
Future work is therefore required to test on a robot platform, removing the assumptions of deterministic action outcomes and perfect sensing.

Additionally, as currently implemented, \alg{} is not optimal in guiding the robot to achieve its goal under the shortest amount of steps. As with FF-Replan, its performance currently cannot be guaranteed; this limitation is amplified in certain domains. For example, \alg{} will not perform well in certain domains that require the robot to plan ahead for contingencies in order to avoid backtracking or dead ends. In future work, we plan to remove this limitation. Once \alg{} has fully built its search tree, it can continue expanding all other nodes to build the equivalent tree of POMCP, thus inheriting its asymptotic optimality guarantees. 




\section{CONCLUSION}
We present \alg{}, a probabilistic planning algorithm that draws inspiration from two existing baselines, POMCP and FF-Replan. \alg{} produces a search tree similar to POMCP, but draws from FF-Replan's determinization step in order to narrow the breadth of search and focus on useful paths. We evaluated \alg{} against both baselines and found that \alg{} often produces plans with a smaller number of steps, is more lightweight than baselines in the domains that we tested it within, and improves with greater amounts of time allocated for planning. In future work, we plan to add functionality to expand \alg{}'s set of nodes so it asymptotically achieves optimality.
\laura{!!! discuss guarantees with FF-Replan in discussion, here, abstract and intro. ALSO should be in results section with experimental results confirming it.}

\bibliographystyle{IEEEtran} 
\bibliography{references}

@inproceedings{NIPS2010_edfbe1af,
 author = {Silver, David and Veness, Joel},
 booktitle = {Adv. Neural Inf. Process. Syst.},
 publisher = {Curran Associates, Inc.},
 title = {Monte-Carlo Planning in Large {POMDPs}},
 year = {2010}
}

@inproceedings{yoon2007ff,
author = {Yoon, Sungwook and Fern, Alan and Givan, Robert},
title = {{FF-Replan}: a baseline for probabilistic planning},
year = {2007},
isbn = {9781577353447},
booktitle = {Proc. Int. Conf. Int. Conf. Automated Planning and Scheduling},
pages = {352–359},
numpages = {8},
location = {Providence, Rhode Island, USA},
}

@inproceedings{little2007probabilistic,
  title={Probabilistic planning vs. replanning},
  author={Little, Iain and Thiebaux, Sylvie and others},
  booktitle={ICAPS Workshop on IPC: Past, Present and Future},
  pages={1--10},
  year={2007}
}

@inproceedings{
paudel2025deploymenttime,
title={Deployment-time Selection of Prompts for {LLM}-informed Object Search in Partially-Known Environments},
author={Abhishek Paudel and Gregory J. Stein},
booktitle={ICRA Workshop on Foundation Models and Neuro-Symbolic AI for Robotics},
year={2025}
}

@article{khanal2025learning,
  title={Learning-Augmented Model-Based Multi-Robot Planning for Time-Critical Search and Inspection Under Uncertainty},
  author={Khanal, Abhish and Mathew, Joseph Prince and Nowzari, Cameron and Stein, Gregory J},
  journal={arXiv:2507.06129},
  year={2025}
}

@ARTICLE{brown2012MCTS,
  author={Browne, Cameron B. and Powley, Edward and Whitehouse, Daniel and Lucas, Simon M. and Cowling, Peter I. and Rohlfshagen, Philipp and Tavener, Stephen and Perez, Diego and Samothrakis, Spyridon and Colton, Simon},
  journal={IEEE Trans. Comput. Intell. AI in Games}, 
  title={A Survey of {M}onte {C}arlo {T}ree {S}earch Methods}, 
  year={2012},
  volume={4},
  number={1},
  pages={1-43},
  doi={10.1109/TCIAIG.2012.2186810}}

@inproceedings{littman1995learning,
  title={Learning policies for partially observable environments: {S}caling up},
  author={Littman, Michael L and Cassandra, Anthony R and Kaelbling, Leslie Pack},
  booktitle={Proc. 12th Int. Conf. Mach. Learn.},
  pages={362--370},
  year={1995},
  publisher={Elsevier}
}

@article{karkus2017qmdp,
  title={{QMDP-Net}: Deep learning for planning under partial observability},
  author={Karkus, Peter and Hsu, David and Lee, Wee Sun},
  journal={Adv. Neural Inf. Process. Syst.},
  volume={30},
  year={2017}
}

@article{somani2013despot,
  title={{DESPOT}: Online {POMDP} planning with regularization},
  author={Somani, Adhiraj and Ye, Nan and Hsu, David and Lee, Wee Sun},
  journal={Adv. Neural Inf. Process. Syst.},
  volume={26},
  year={2013}
}

@article{fox2003pddl2,
  title={{PDDL2.1}: An extension to {PDDL} for expressing temporal planning domains},
  author={Fox, Maria and Long, Derek},
  journal={J. Artif. Intell. Res.},
  volume={20},
  pages={61--124},
  year={2003}
}

@article{richter2010lama,
  title={The {LAMA} planner: Guiding cost-based anytime planning with landmarks},
  author={Richter, Silvia and Westphal, Matthias},
  journal={J. Artif. Intell. Res.},
  volume={39},
  pages={127--177},
  year={2010}
}

@inproceedings{coulom2006efficient, 
author = {Coulom, R\'{e}mi}, 
title = {Efficient selectivity and backup operators in Monte-Carlo tree search}, year = {2006}, isbn = {3540755373}, 
publisher = {Springer-Verlag}, 
booktitle = {Proc. 5th Int. Conf. Computers and Games}, 
pages = {72–83}, 
numpages = {12}, 
location = {Turin, Italy}}

@article{KAELBLING199899,
title = {Planning and acting in partially observable stochastic domains},
journal = {Artif. Intell.},
volume = {101},
number = {1},
pages = {99-134},
year = {1998},
issn = {0004-3702},
author = {Leslie Pack Kaelbling and Michael L. Littman and Anthony R. Cassandra}
}

@inproceedings{sunberg2018online,
  title={Online algorithms for {POMDPs} with continuous state, action, and observation spaces},
  author={Sunberg, Zachary and Kochenderfer, Mykel},
  booktitle={Proc. Int. Conf. Automated Planning and Scheduling},
  pages={259--263},
  year={2018}
}

@INPROCEEDINGS{curtis2023task,
  author={Curtis, Aidan and Kaelbling, Leslie and Jain, Siddarth},
  booktitle={Proc. IEEE Int. Conf. Robot. Autom.}, 
  title={Task-Directed Exploration in Continuous {POMDPs} for Robotic Manipulation of Articulated Objects}, 
  year={2023},
  pages={3721-3728}
  }

@article{helmert2006fast,
  title={The fast downward planning system},
  author={Helmert, Malte},
  journal={J. Artif. Intell. Res.},
  volume={26},
  pages={191--246},
  year={2006}
}

@inproceedings{couetoux2011continuous,
  title={Continuous upper confidence trees},
  author={Cou{\"e}toux, Adrien and Hoock, Jean-Baptiste and Sokolovska, Nataliya and Teytaud, Olivier and Bonnard, Nicolas},
  booktitle={Int. Conf. Learn. Intell. Optim.},
  pages={433--445},
  year={2011},
  publisher={Springer}
}

@inproceedings{kim2019pomhdp,
  title={{POMHDP}: Search-based belief space planning using multiple heuristics},
  author={Kim, Sung-Kyun and Salzman, Oren and Likhachev, Maxim},
  booktitle={Proc. Int. Conf. Automated Planning and Scheduling},
  pages={734--744},
  year={2019}
}

@article{rhodes2023autonomous,
  title={Autonomous search of an airborne release in urban environments using informed tree planning},
  author={Rhodes, Callum and Liu, Cunjia and Westoby, Paul and Chen, Wen-Hua},
  journal={Autonomous Robots},
  volume={47},
  number={1},
  pages={1--18},
  year={2023},
  publisher={Springer}
}

@inproceedings{ramrakhya2022habitat,
  title={Habitat-web: Learning embodied object-search strategies from human demonstrations at scale},
  author={Ramrakhya, Ram and Undersander, Eric and Batra, Dhruv and Das, Abhishek},
  booktitle={Proc. IEEE/CVF Conf. Comput. Vis. and Pattern Recognition},
  pages={5173--5183},
  year={2022}
}

@article{chiou2022towards,
  title={Towards human--robot teaming: Tradeoffs of explanation-based communication strategies in a virtual search and rescue task},
  author={Chiou, Erin K and Demir, Mustafa and Buchanan, Verica and Corral, Christopher C and Endsley, Mica R and Lematta, Glenn J and Cooke, Nancy J and McNeese, Nathan J},
  journal={Int. J. Social Robot.},
  volume={14},
  number={5},
  pages={1117--1136},
  year={2022},
  publisher={Springer}
}


\end{document}